\documentclass[11pt]{article} 
\usepackage{rldm,palatino}
\usepackage{graphicx}
\usepackage[colorlinks=true,linkcolor=blue,citecolor=blue,urlcolor=blue]{hyperref}
\usepackage{amscd,amsfonts,amssymb,amstext,latexsym} 
\usepackage{amsmath,mathbbol,mathrsfs,stmaryrd, mathtools} 
\usepackage[numbers,sort&compress]{natbib}
\usepackage{caption}
\usepackage{subcaption}
\usepackage{algorithm}
\usepackage{algpseudocode}
\usepackage{todonotes}
\usepackage[colorlinks=true,linkcolor=blue,citecolor=blue,urlcolor=blue]{hyperref}

\usepackage{wrapfig}

\usepackage{graphicx}
\usepackage{amscd,amsfonts,amssymb,amstext,latexsym} 
\usepackage{amsmath,mathbbol,mathrsfs,stmaryrd, mathtools} 
\usepackage[numbers,sort&compress]{natbib}
\usepackage{caption}
\usepackage{subcaption}
\usepackage{algorithm}
\usepackage{algpseudocode}
\usepackage{todonotes}

\usepackage{graphicx,wrapfig,adjustbox,lipsum}
\usepackage[inline]{enumitem}
\usepackage{amsthm}

\usepackage{mathtools}

\DeclarePairedDelimiterX{\setarg}[1]{\{}{\}}{%
  \ifnum\currentgrouptype=16 \else\begingroup\fi
  \activatebar#1
  \ifnum\currentgrouptype=16 \else\endgroup\fi
}

\DeclarePairedDelimiterX{\expectarg}[1]{[}{]}{%
  \ifnum\currentgrouptype=16 \else\begingroup\fi
  \activatebar#1
  \ifnum\currentgrouptype=16 \else\endgroup\fi
}

\newcommand{\prob}{\operatorname{\Pr}\probarg}
\DeclarePairedDelimiterX{\probarg}[1]{(}{)}{%
  \ifnum\currentgrouptype=16 \else\begingroup\fi
  \activatebar#1
  \ifnum\currentgrouptype=16 \else\endgroup\fi
}

\newcommand{\innermid}{\nonscript\;\delimsize\vert\nonscript\;}
\newcommand{\activatebar}{%
  \begingroup\lccode`\~=`\|
  \lowercase{\endgroup\let~}\innermid 
  \mathcode`|=\string"8000
}

\title{Learning Abstract and Transferable Representations for Planning}
\author{
Steven James, Benjamin Rosman \\
School of Computer Science and Applied Mathematics\\
University of the Witwatersrand\\
Johannesburg, South Africa \\
\texttt{\{steven.james,benjamin.rosman1\}@wits.ac.za} \\
\And
George Konidaris \\
Department of Computer Science\\
Brown University\\
Providence RI, 02912 \\
\texttt{gdk@cs.brown.edu}
}

\begin{document}

\maketitle

\begin{abstract}

We are concerned with the question of how an agent can acquire
its own representations from sensory data. We restrict our focus to learning representations
for long-term planning, a class of problems that state-of-the-art learning
methods are unable to solve. 
We propose a framework for autonomously learning state abstractions
of an agent’s environment, given a set of skills. Importantly, these abstractions are task-independent,
and so can be reused to solve new tasks.
We demonstrate how an agent can use an existing set of options
to acquire representations from ego- and object-centric observations. 
These abstractions can immediately be reused by the same agent in new environments. We show how to combine these portable representations with problem-specific ones to generate a sound description of a specific task that can be used for abstract planning. 
Finally, we show how to autonomously construct a multi-level hierarchy consisting
of increasingly abstract representations. Since these hierarchies are transferable,
higher-order concepts can be reused in new tasks, relieving the agent from relearning
them and improving sample efficiency. 
Our results demonstrate that our approach allows an agent
to transfer previous knowledge to new tasks, improving sample efficiency as the
number of tasks increases.

\end{abstract}

\keywords{
representation learning, planning, abstractions, hierarchy
}


\startmain 

\section{Introduction}

Recently, state-of-the-art reinforcement learning (RL) approaches have made several significant advances in challenging domains, such as controlling nuclear reactors \citep{degrave_magnetic_2022}.
Despite these successes, it is clear that these approaches do not capture a remarkable aspect of human intelligence---namely, that humans can solve not just a single problem, but a massively diverse array of tasks.
Consider the \textsc{AlphaZero} agent which attained superhuman performance in the grand challenge of Go \citep{silver16}.
While this is an immensely difficult task, the input to this agent is a set of binary vectors specifying stone locations, while the output is a location at which to place a stone.
This input-output format is provided to the agent by a human designer because it captures exactly the task that must be accomplished. 
However, it means that the agent cannot solve any other tasks by definition; it cannot drive a car or cook a meal. While the former approach is useful in designing narrow, application-specific solutions, it falls short of the ultimate aim of generally intelligent agents.

In general, tasks in RL are formulated by human designers and provided to agents in a standardised, compact form. 
Though this practice is widespread, it sidesteps an important question: \textit{where do these representations come from in the first place?}
It is obvious that this approach is infeasible in the long run: we cannot preprogram
an agent with every task it may encounter before deploying it in the real world. Nor can we require that a human designer accompany the agent throughout its lifetime, providing task representations as and when required. Clearly then, the only option is for the
agent to learn its own representation for any newly encountered task directly from its observations of the world.

If we are to design a \emph{single} agent capable of solving multiple tasks in the real world, it must necessarily have a complex sensorimotor space.
However, solving long-horizon tasks at this low level is typically infeasible.
A common approach to tackling this problem is hierarchical RL, which makes use of \textit{abstractions} to simplify the problem. 
Action abstractions (also known as \textit{skills}) alleviates the need to reason using low-level actions, while the use of state abstraction (where states are aggregated into high-level states) reduces the size of the problem.
However, if an agent’s abstractions are too high-level, it risks omitting important and necessary
details. Conversely, if it seeks to preserve every last detail of the environment, then its
representations will be too low-level and planning will once again be infeasible. The
key question is how best to construct an abstract model of an environment while
retaining only the information required for planning.

In this work, we outline a framework for learning transferable abstract representations from low-level data that can be used for long-term planning. More concretely, we extend the framework of \citet{konidaris18} so that the learned representations are portable---given a new task, an agent can reuse
the representations it has learned previously to speed up learning.
We apply our framework to learn both agent- and object-centric representations in several high-dimensional domains, and demonstrate that our approach results in agents that are i) more sample efficient; ii) able to learn their own representations; and iii) able to use their learned representations to solve a variety of tasks. 

\section{Preliminaries} \label{sec:reps}

We begin by assuming that an agent is equipped with a set of skills and model tasks as semi-Markov decision processes $\mathcal{M} = \langle \mathcal{S}, \mathcal{O}, \mathcal{T}, \mathcal{R} \rangle$ where 
\begin{enumerate*}[label=(\roman*)]
\item $\mathcal{S}$ is the state space;
\item $\mathcal{O}(s)$ is the set of temporally-extended actions known as \textit{options} available at state $s$;
\item $\mathcal{T}$ describes the transition dynamics, specifying the probability of arriving in state $s^\prime$ after option $o$ is executed from $s$; and
\item $\mathcal{R}$ specifies the reward for reaching state $s'$ after executing option $o$ in state $s$. 
\end{enumerate*}
An option $o$ is defined by the tuple $\langle I_o, \pi_o; \beta_o \rangle$, where $I_o$ is the
\textit{initiation set} specifying the states where the option can be executed, $\pi_o$ is the \textit{option policy} which specifies the actions to execute, and $\beta_o$ the probability of the option terminating  in each state \citep{sutton99}.


We intend to learn an abstract representation suitable for planning. Prior work has shown that a sound and complete abstract representation must necessarily be able to estimate the set of initiating and terminating states for each option \citep{konidaris18}. 
In classical planning, this corresponds to the \textit{precondition} and \textit{effect} of each high-level action operator.
The precondition is defined as $\text{Pre}(o) = \prob{s \in I_o}$, which is a probabilistic classifier that expresses the probability that option $o$ can be executed at state $s$.
Similarly, the effect represents the distribution of states an agent may find itself in after executing an option from states drawn from some starting distribution \citep{konidaris18}.
Since the precondition is a probabilistic classifier and the effect is a density estimator, they can be learned directly from option execution data. 
We can use  preconditions and effects to evaluate the probability of an arbitrary sequence of options---a plan---executing successfully.

\paragraph{Partitioned Options}

For large or continuous state spaces, estimating $\prob{s' | s, o}$ is difficult; however, if we assume that terminating states are independent of starting states, we can make the simplification  $\prob{s' | s, o} = \prob{s' | o}$. 
These \textit{subgoal} options are not overly restrictive, since they refer to options that drive an agent to some set of states with high reliability.
While many options are not subgoal, it is often possible to \textit{partition} an option's initiation set into a finite number of subsets. 
That is, we partition an option $o$'s start states into finite regions $\mathcal{C}$ such that $\prob{s^\prime | s, o, c} \approx \prob{s^\prime | o, c}, c \in \mathcal{C}$. 
Given (partitioned) subgoal options, we can estimate their preconditions and effects using the approach outlined by \citet{konidaris18}. 

\section{Agent-Centric Abstractions} \label{sec:agent-learning}

Central to the field of artificial intelligence is the notion of the \textit{agent}. 
Real-world agents are robots, which perceive their environments through sensors and act upon them with effectors.
In practice, a human designer will usually build upon the observations produced by the agent’s sensors to construct the Markov state space for the problem at hand, while discarding unnecessary perceptual information. Instead we will seek
to effect transfer by using both the agent’s sensor information---which is typically egocentric---in addition to the Markov state space.
We assume that tasks are related because they are faced by the same agent. For example, consider a robot (equipped with various sensors) that is required to perform a number of as yet unspecified tasks. The only aspect that remains constant across all these tasks is the presence of the robot and, more importantly, its sensors, which map the state space $\mathcal{S}$ to a portable, lossy and egocentric observation space $\mathcal{D}$ known as \textit{agent space}.  
 We can use $\mathcal{D}$ to define portable options, whose option policies, initiation sets and termination conditions are all defined egocentrically. Because $\mathcal{D}$ remains constant regardless of the underlying SMDP, these options can be transferred across tasks \citep{konidaris07}.
 
\begin{figure}[h!]
\begin{minipage}[b]{.5\textwidth}
\centering
\includegraphics[width=1\textwidth]{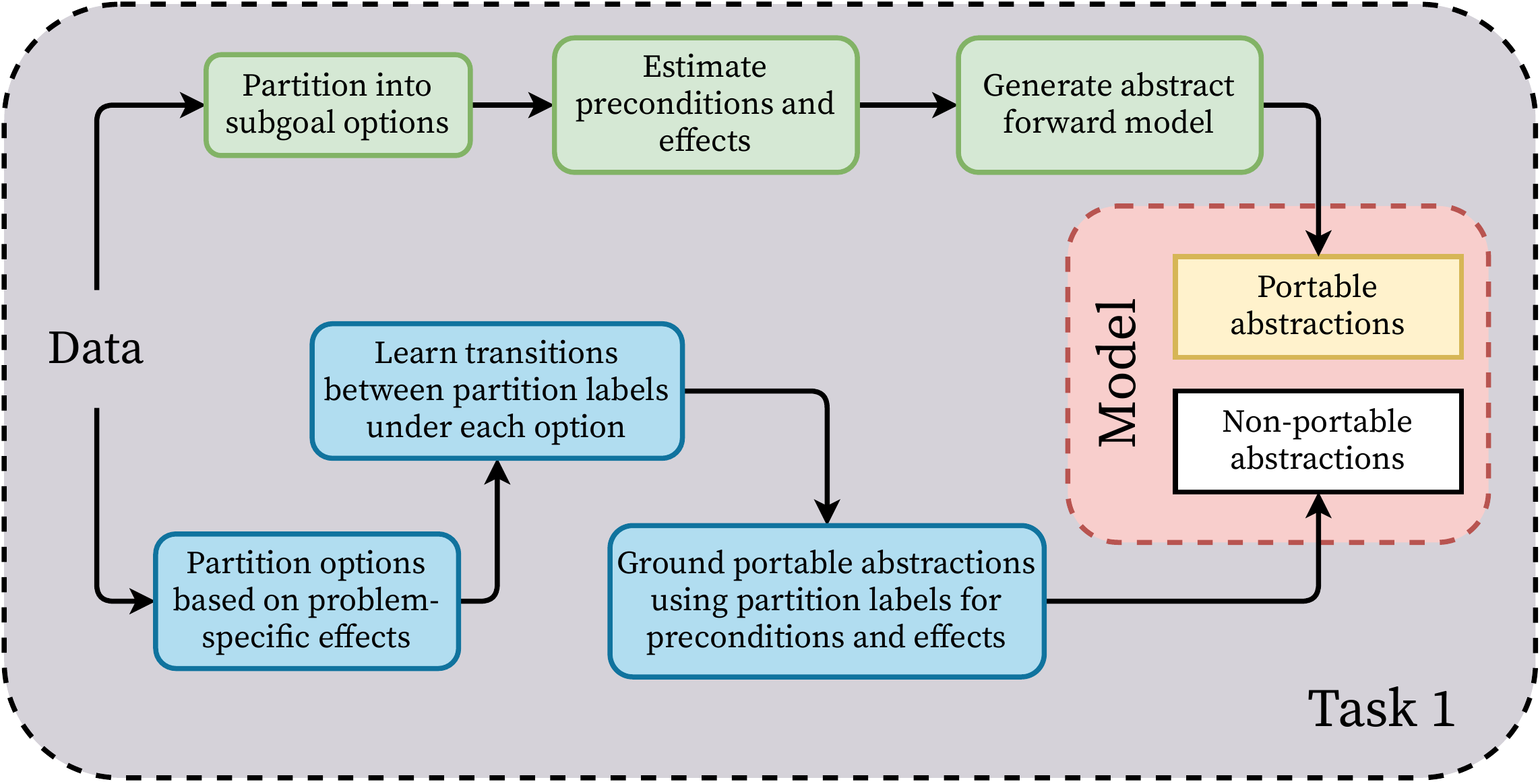}
	\caption{The agent learns transferable representations, which are then combined with problem-specific abstractions to form a model suitable for planning. }
	\label{fig:overview}      
\end{minipage}
\hfill
\begin{minipage}[b]{.45\textwidth}
\centering
\includegraphics[width=0.85\textwidth]{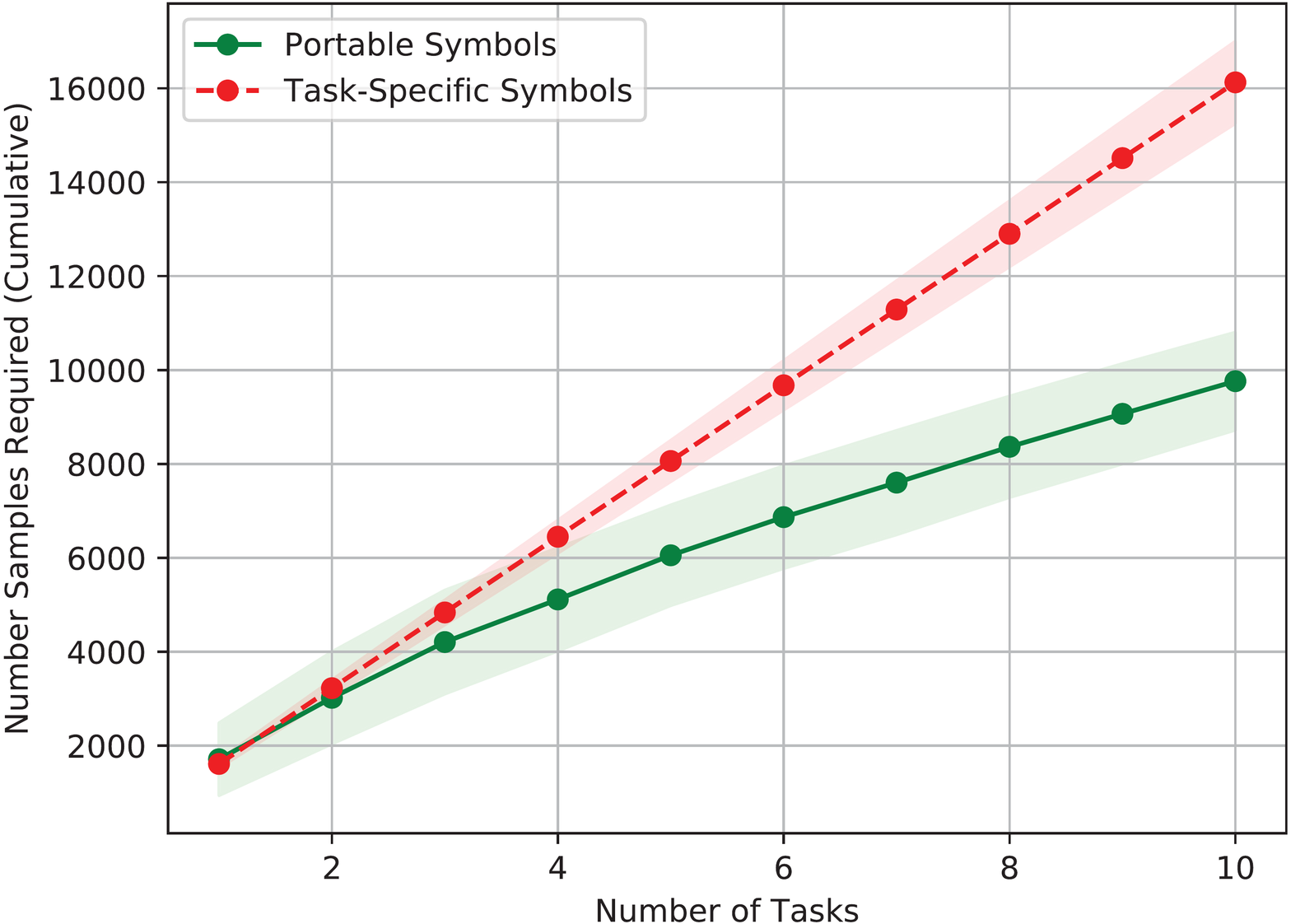}
\caption{Owing to transfer, the number of samples required by the agent to learn a sufficiently accurate model decreases with the number of tasks faced.} \label{fig:results}
\end{minipage}
\end{figure}

Having made this distinction, we can write the state space of any given task $\mathcal{M}_i$ as the tuple $\langle \mathcal{X}_i, \mathcal{D} \rangle$, where $\mathcal{D}$ is shared across tasks and $\mathcal{X}_i$ represents task-specific state variables. 
Given this representation, we can follow a two-step process. The first phase uses the procedure outlined in Section~\ref{sec:reps} to learn portable abstract rules using agent-space transition data only. 
The second phase uses problem-space transitions to partition options in $\mathcal{X}_i$. 
Each partition is assigned a unique label, and these labels are used as parameters to ground the previously learned portable representations in the current task.
For a new task, the agent need only estimate how the partition labels change under each option execution. 
Figure~\ref{fig:overview} illustrates this entire process, but see \citet{james20} for more details.


We test our approach in the \textit{Treasure Game} \citep{konidaris18}, where an agent navigates a maze in search of treasure. 
This domain contains ladders and doors which impede the agent. Some doors can be opened and closed with levers, while others require a key to unlock.
We first learn an abstract representation using agent-space transitions only, following the same procedure above.
Once we have learned sufficiently accurate portable abstractions, they need only be instantiated for the given task by learning the linking between partitions.
This requires far fewer samples than learning a task-specific representation from scratch.
To illustrate, we construct ten levels and gather transition samples from each task. We use these samples to build both task-specific and  egocentric (portable) models.
For each level, we collect data until a model is sufficiently accurate
at which point we continue to the next task. Results are given by Figure~\ref{fig:results}.

\section{Object-Centric Abstractions} \label{sec:minecraft}

Having assumed the existence of an agent, it is natural to make another assumption---that the world consist of objects, and that similar objects are common amongst tasks.
Previously, we assumed the existence of an agent equipped with sensors, which led to the idea of agent space. 
Since we are now assuming the existence of objects, a natural extension is to introduce the notion of \textit{object space}. 
We adopt an object-centric formulation: in a task with $n$ objects, the state is represented by the set 
$\{ \mathbf{f}_{a}, \mathbf{f}_{1}, \mathbf{f}_{2}, \ldots,  \mathbf{f}_{n}  \},$ where $\mathbf{f}_{a}$ is a vector of the agent's features and $\mathbf{f}_{i}$ are the features of object $i$  \citep{ugur15}.

\begin{figure}[b!]
\begin{minipage}[b]{.45\textwidth}
\centering
\includegraphics[width=.95\textwidth]{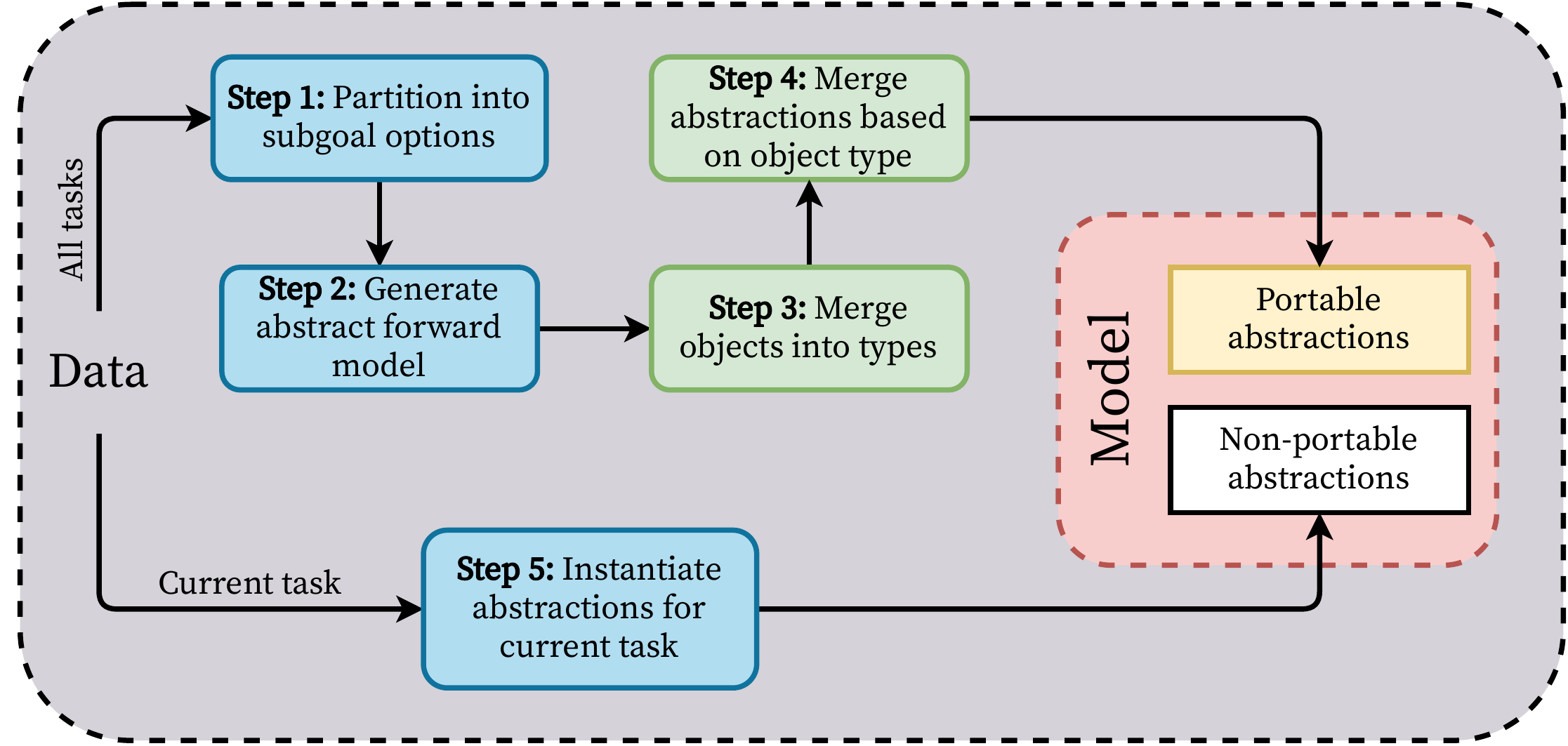} 
\caption{Learning object-relative representations from data. Blue nodes represent problem-specific representations, while green nodes are abstractions that can be transferred between tasks.} \label{fig:object-process2}
\end{minipage}
\hfill
\begin{minipage}[b]{.45\textwidth}
\centering
\includegraphics[width=0.65\textwidth]{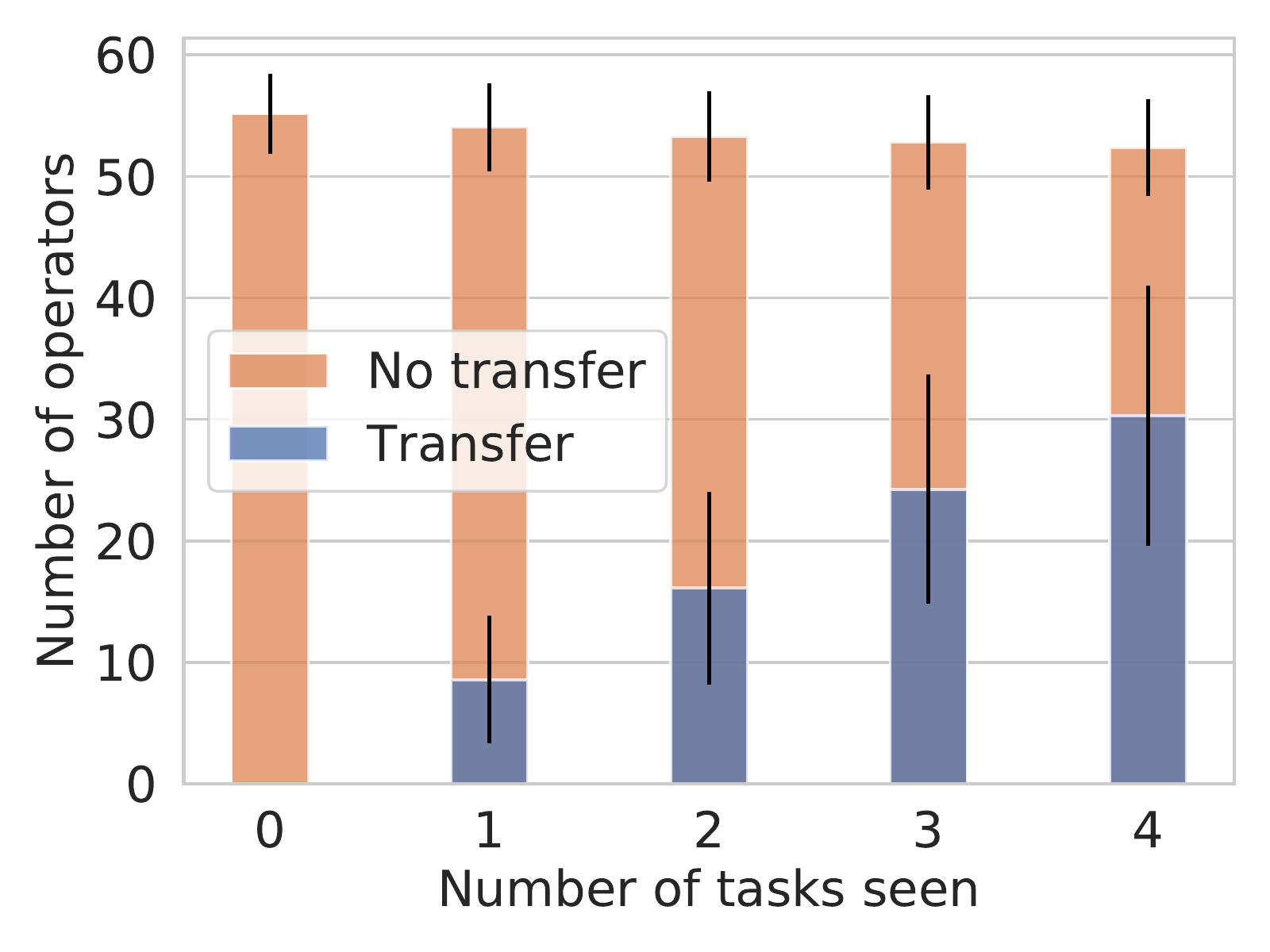}
\caption{Orange: number of operators that must be learned to produce a sufficiently accurate model of a task. Blue: number of operators transferred between tasks.  Mean and standard deviation over 80 runs.}\label{fig:result2}
\end{minipage}
\end{figure}

The process to learn a grounded representation is now three-fold and is summarised by Figure~\ref{fig:object-process2}.
We first follow the same procedure outlined in Section~\ref{sec:reps}  to construct a non-portable representation of a single task.
Since object space is already factored into the constituent objects, each abstraction will refer to a distribution over a particular object's state. 
Next, we merge these representations where objects fall into the same ``type'' using the notion of \textit{effect equivalence} \citep{sahin07}---two objects are grouped into the same type when they undergo similar effects under the same set of options.
Finally, we once more use the problem-specific state data to construct partition labels, which are used to  ground previously learned portable representations in the current task.
See \citet{james22} for more details.

\begin{wrapfigure}{r}{0.4\textwidth}
  \vspace{-15pt}
    \centering
        \begin{subfigure}[t]{0.3\linewidth}
    	\centering
        \includegraphics[width=\linewidth]{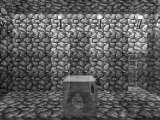}
        \caption{\texttt{rep\_15}}
    \end{subfigure}
    \begin{subfigure}[t]{0.3\linewidth}
    	\centering
        \includegraphics[width=\linewidth]{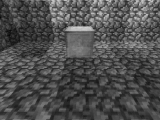}
        \caption{\texttt{rep\_2}}
    \end{subfigure}
    \quad
    \begin{subfigure}[t]{0.33\linewidth}
    	\centering
        \includegraphics[height=20mm]{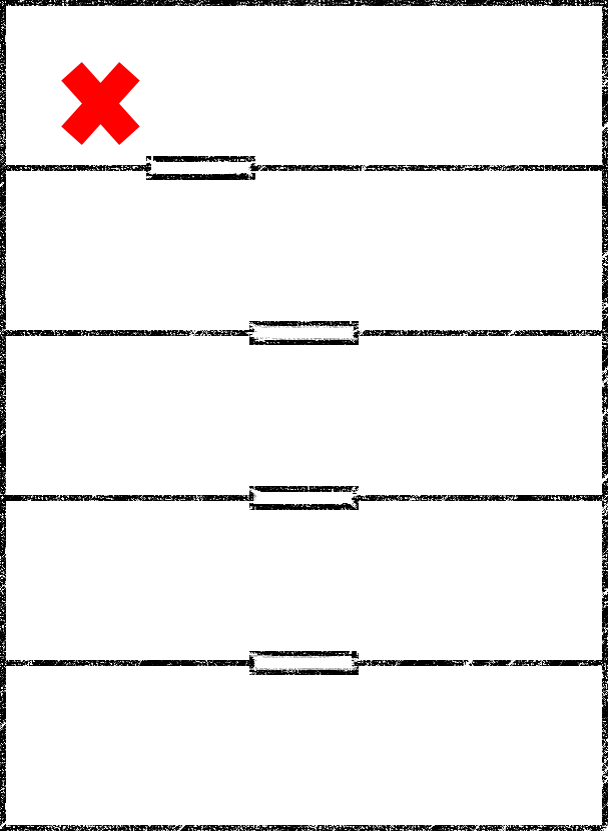}
        \caption{\texttt{{\color{red}rep\_17}}}
    \end{subfigure}
    \par\bigskip
    \begin{subfigure}[t]{0.3\linewidth}
    	\centering
        \includegraphics[width=\linewidth]{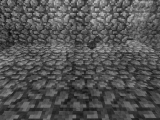}
        \caption{\texttt{rep\_19}}
    \end{subfigure}
    \begin{subfigure}[t]{0.3\linewidth}
    	\centering
        \includegraphics[width=\linewidth]{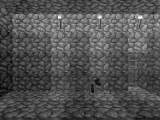}
        \caption{\texttt{rep\_20}}
    \end{subfigure}
\caption{Abstract precondition and effect for breaking a gold block. The agent must be standing in front of a gold block  (\texttt{rep\_15}) at a particular location (\texttt{rep\_17}), and the gold block must be whole (\texttt{rep\_2}). As a result, the agent finds itself in front of a disintegrated block (\texttt{rep\_20}), and the gold block is disintegrated (\texttt{rep\_19}). Only the red abstraction must be relearned for each new task.}     \label{fig:attack-block} 
   \vspace{-25pt}
\end{wrapfigure}


We demonstrate our approach in a series of Minecraft levels, where each consists of five rooms with various items positioned throughout.
Rooms are connected with either regular doors, which can be opened by direct interaction, or  puzzle doors requiring the agent to pull a lever to open. 
The world is described by the state of each of the objects (given directly by each object's appearance as a $600 \times 800$ RGB image), the agent's view, and current inventory.
The agent is given high-level skills, such as \texttt{ToggleDoor} and \texttt{WalkToItem}.
To simplify learning, we downscale images and applying PCA to a greyscaled version, preserving the top 40 principal components.
We follow the process in Figure~\ref{fig:object-process2} to learn portable object-centric representations, and ground them with task-specific partition labels derived from the agent's $xyz$-location.
As mentioned, objects are grouped into types based on their effects, which is made easier because certain objects do not undergo effects under certain options. 
For example, the chest cannot be toggled, while a door can, and thus it is immediately clear that they are not of the same type. 
We investigate transferring abstractions between five procedurally-generated tasks, where each task differs in the location of the objects and doors.
For a given task, the agent transfers all operators learned from previous tasks, and continues to collect samples using uniform random exploration until it produces a model that predicts the optimal plan can be executed.
Figure~\ref{fig:attack-block} illustrates a learned abstraction, while Figure~\ref{fig:result2} shows the number of abstract option representations (operators) transferred between tasks.

\section{Hierarchies of Abstractions}

The previous approaches, whether agent- or object-centric, resulted in an abstract decision problem. 
If we apply our framework repeatedly, it will discover increasingly higher order representations, which are themselves distributions over the representations at the level below;  in the agent-centric setting, an abstract state space at level $i > 0$ is the tuple $\langle \mathcal{X}^{(i)}, \mathcal{D}^{(i)} \rangle$, where each $x \in \mathcal{X}^{(i)}$ and $d \in \mathcal{D}^{(i)}$ is a distribution over states in $\mathcal{X}^{(i-1)}$ and $\mathcal{D}^{(i-1)}$ respectively.
The above formulation means that, as we construct more levels in the hierarchy, the resulting representations become increasingly compact and faster to plan with, but so does the degree of uncertainty. 


Our first step is to construct an abstract representation using the approach in Section~\ref{sec:agent-learning}.
Next we must decide how best to discover higher order skills in this new representation.  
We achieve this by converting our representation to a transition graph, identify ``important'' nodes (using the \textsc{VoteRank} metric), and then construct options to reach these \textit{subgoal} nodes using Djikstra's algorithm.
Edges along these paths constitute our higher-order options.
Note that since all the options contain only a single node in their termination set, they are subgoal by construction.   
We can then simply iterate this approach to construct an entire abstraction hierarchy.

\begin{figure}[h!]
\begin{minipage}[b]{.5\textwidth}
\centering
\includegraphics[width=0.8\textwidth]{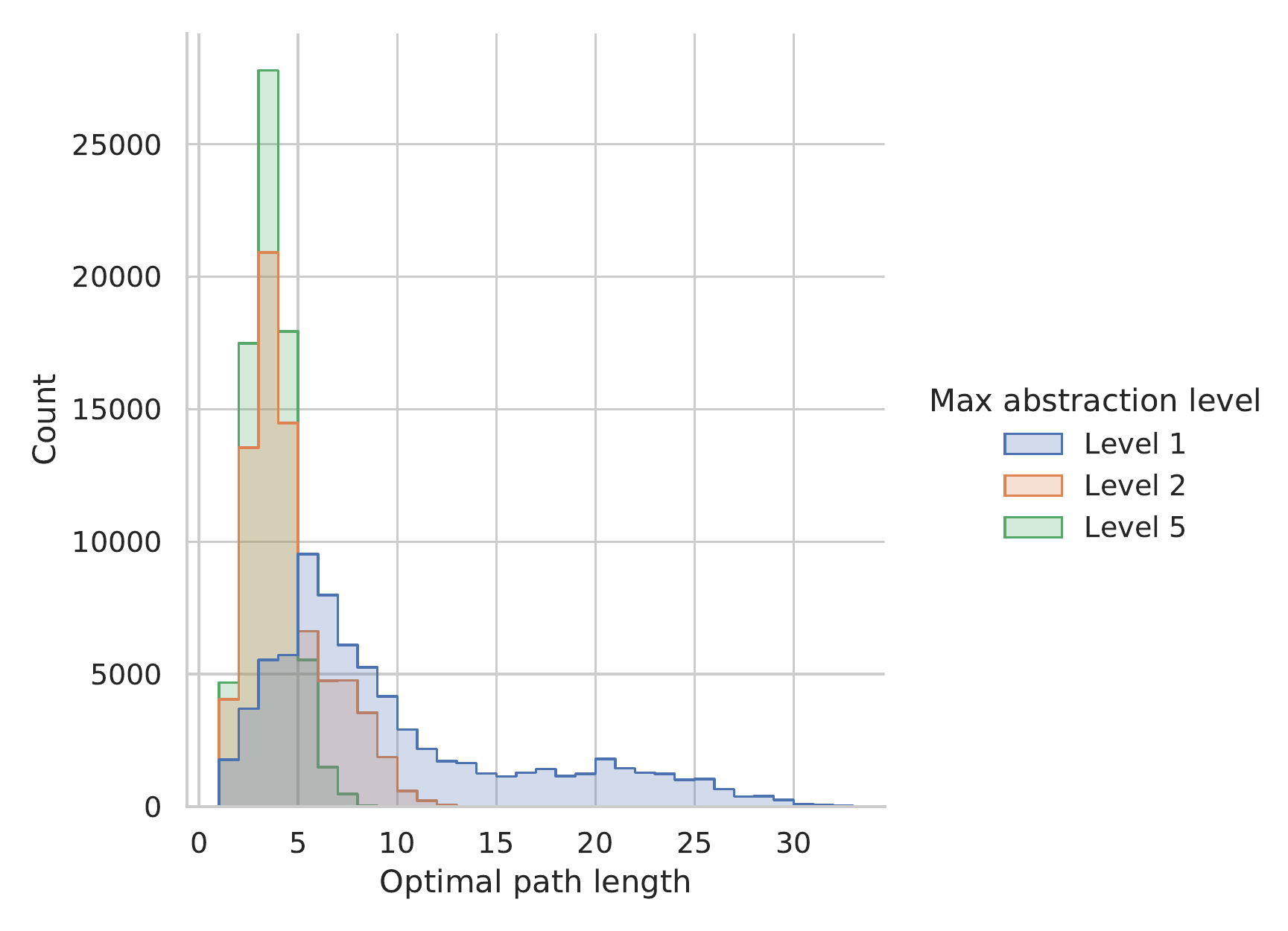}
\caption{Distribution of optimal plan lengths in the first task when using hierarchies of varying heights.} \label{fig:histos}
\end{minipage}
\hfill
\begin{minipage}[b]{.45\textwidth}
\centering
\includegraphics[width=0.8\textwidth]{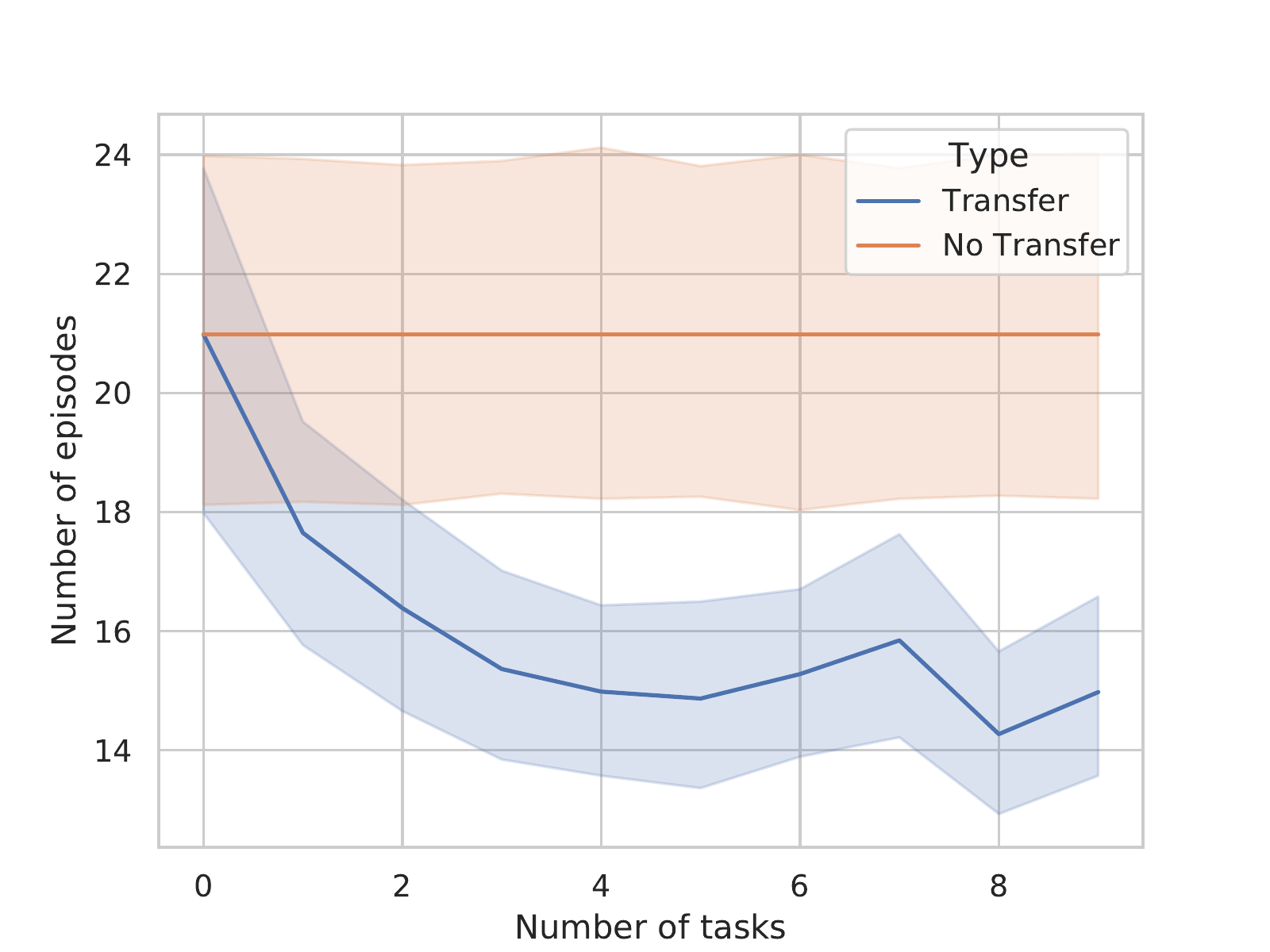}
        \caption{Number of episodes required to learn a model of a given task, decreasing as the agent observes more tasks. Mean and variance reported over 100 runs.}  \label{fig:sample-efficiency}     
\end{minipage}
\end{figure}

We again apply our approach to the \textit{Treasure Game} to construct portable hierarchies.
To illustrate the effect of the hierarchy, we compute the distribution of the length of all pairs of shortest paths for each of the tasks when using abstractions from varying levels of the hierarchy.
Results for the first task are given by Figure~\ref{fig:histos} and indicate that incorporating information at increasingly abstract levels of the hierarchy reduces the size of the graph (this trend holds across all other levels too). 
Consequently, the maximum planning horizon is shortened, which greatly simplifies the planning problem.
We also investigate transfer by presenting the agent with each of the ten tasks in sequence.
Unlike previously, portable representations here consist of representations at various levels in the hierarchy. 
We measure the number of samples required to learn a model of a new task, with the results illustrated by Figure~\ref{fig:sample-efficiency}.
Although the results exhibit high variance (due to the exploration strategy, the differences in tasks, and the randomised task order), sample efficiency is clearly improved when an agent is able to reuse past knowledge.

\section{Conclusion}

We proposed a framework for autonomously learning reusable representations. We showed how to learn an agent- and object-centric representation that can be used for planning. These representations can be transferred to new tasks, reducing the number of times an agent is required to interact with the world. We also showed how to construct a portable hierarchy of abstractions
that can be used to plan at different levels. Altogether, our results indicate
that the learned abstractions can be reused in new tasks, reducing the number of
times an agent is required to interact with its environment. We believe this will be critical to scaling abstraction learning approaches to real
world tasks in the future.

{\footnotesize
\bibliography{main}
}
\end{document}